\documentclass{article}

\usepackage[numbers]{natbib}

\usepackage[preprint]{neurips_2020}
\usepackage{float} 
\usepackage{amsmath} 
\usepackage{booktabs} 
\usepackage{amssymb} 
\usepackage[utf8]{inputenc} 
\usepackage[T1]{fontenc}    
\usepackage{hyperref}       
\usepackage{url}            
\usepackage{amsfonts}       
\usepackage{nicefrac}       
\usepackage{microtype}      
\usepackage{graphicx}       
\usepackage{adjustbox} 
\usepackage{algorithm}
\usepackage{algorithmic}
\makeatletter
\renewcommand{\@noticestring}{} 
\makeatother

\title{Course Project Report: Comparing MCMC and Variational Inference for Bayesian Probabilistic Matrix Factorization on the MovieLens Dataset}

\author{%
  Ruixuan Xu \\
  Department of Computer Science and Engineering\\
  The Chinese University of Hong Kong\\
  \texttt{1155211213@link.cuhk.edu.hk} \\
  \And
  Xiangxiang Weng \\
  Department of Computer Science and Engineering\\
  The Chinese University of Hong Kong\\
  \texttt{1155211173@link.cuhk.edu.hk} \\
}

\begin{document}

\maketitle

\begin{abstract}
This is a course project report with complete methodology, experiments, references and mathematical derivations. Matrix factorization \cite{1} is a widely used technique in recommendation systems. Probabilistic Matrix Factorization (PMF) \cite{2} extends traditional matrix factorization by incorporating probability distributions over latent factors, allowing for uncertainty quantification. However, computing the posterior distribution is intractable due to the high-dimensional integral. To address this, we employ two Bayesian inference methods: Markov Chain Monte Carlo (MCMC) \cite{3,4} and Variational Inference (VI) \cite{5,6} to approximate the posterior. We evaluate their performance on MovieLens dataset \cite{7} and compare their convergence speed, predictive accuracy, and computational efficiency. Experimental results demonstrate that VI offers faster convergence, while MCMC provides more accurate posterior estimates.
\end{abstract}

\section{Introduction}

Collaborative filtering \cite{8} is an essential technique in recommendation systems, where the goal is to predict user preferences based on sparse observed ratings. Matrix factorization \cite{1} has been widely adopted due to its ability to model user-item interactions effectively. However, traditional matrix factorization relies on point estimates, which may lead to overfitting and lack of uncertainty quantification.

Probabilistic Matrix Factorization (PMF) \cite{2} mitigates this issue by considering a probabilistic approach, where user and item latent matrices are treated as random variables with prior distributions. The challenge in PMF is computing the posterior distribution of latent matrices, which is intractable. To approximate the posterior, we explore two Bayesian inference methods:

1. Markov Chain Monte Carlo (MCMC) \cite{3,4}: A sampling-based approach that provides asymptotically exact posterior estimates.

2. Variational Inference (VI) \cite{5,6}: An optimization-based approach that approximates the posterior using a parameterized distribution.

In this report, we implement both methods on MovieLens dataset \cite{7} and compare their performance.

\section{Problem Setting}

\subsection{Matrix Factorization Model}

In mathematics, a sparse matrix refers to a matrix in which most of the elements are zero. A sparse rating matrix is a common data structure in recommendation systems, typically used to represent users' rating data for items. Collaborative filtering \cite{8} relies on user behavior data such as ratings to predict items a user might like. Due to the high sparsity of rating matrices, directly storing and computing them would waste a lot of memory and computational resources. Matrix factorization \cite{1} addresses this by decomposing the user-item rating matrix into two low-dimensional matrices, and then predicting ratings via their dot product, thereby achieving collaborative filtering.

Given an $N \times M$ sparse rating matrix $R$. Suppose there are $L$ observed ratings ($L \ll N \times M$), for predicting all unrated items, we need to estimate $N \times M - L \approx N \times M$ parameters. An $N\times M$ rating matrix $R$ can be approximated by the multiplication of an $N \times K$ matrix and a $K \times M$ matrix. Here, $K$ is feature dimension, $U \in \mathbb{R}^{N\times K}$ is user feature matrix, and $V \in \mathbb{R}^{M \times K}$ is item feature matrix. With matrix factorization, rather than estimating $N \times M - L \approx N \times M$ parameters, we only need to estimate $(N+M) \times K$, where $K \ll \min(N,M)$. Our goal is to find matrices $U$ and $V$ such that $UV^T$ can accurately represent the original rating matrix. In this way, the number of parameters to be estimated is greatly reduced, and the entire prediction matrix can be obtained simply by matrix multiplication, making computation efficient. Finding the target parameter matrices $U$ and $V$ involves optimization methods based on gradient descent. For example, minimizing the following loss function:
$$
\min\limits_{U,V} \sum\limits_{(i,j)\in O}(r_{ij}-\bold{u}_i\bold{v}_j^T)^2
$$
where:

- $O$ is the set of all rated items.

- $r_{ij}$ is the true rating given by user $i$ to item $j$.

- $\bold{u}_i\bold{v}_j^T$ is the predicted rating.

Gradient descent update rules:
\begin{align*}
\bold{u}_i &\leftarrow \bold{u}_i+ \alpha \cdot \sum\limits_{j\in O_i}(r_{ij}-\bold{u}_i\bold{v}_j^T)\bold{v}_j\\
\bold{v}_j &\leftarrow \bold{v}_j+ \alpha \cdot \sum\limits_{i\in O_j}(r_{ij}-\bold{u}_i\bold{v}_j^T)\bold{u}_i
\end{align*}
where $\alpha$ is the learning rate.

However, traditional matrix factorization has many limitations, such as poor generalization ability, inability to capture the uncertainty of latent vectors and lack of probabilistic interpretation. Therefore, we adopt Bayesian Probabilistic Matrix Factorization (BPMF) \cite{4,6} on the basis of matrix factorization.

\subsection{Bayesian Probabilistic Matrix Factorization}

Instead of estimating matrices $U$ and $V$ to compute each rating $r_{ij}$ based on deterministic approaches, one can use Bayesian inference. Consider each row of $U$ and $V$, i.e., $\bold{U}_i$ and $\bold{V}_j$, as a multivariate random variables. $\bold{U}_i$ and $\bold{V}_j$ are assumed to be standard normal random vectors.
\begin{align*}
\begin{cases}
U_{ik} \sim N(0,1) \quad \text{ for any } 1 \le k \le K \\
V_{ik} \sim N(0,1) \quad \text{ for any } 1 \le k \le K 
\end{cases}
\end{align*}
The prior distributions of $\bold{U}_i$ and $\bold{V}_j$ have the following PDFs
\begin{align*}
\begin{cases}
f_{\bold{U}_{i}}(\bold{u}_i)=\frac{1}{(2\pi)^{\frac{K}{2}}} \exp\left(-\frac{1}{2}\bold{u}_i\bold{u}_i^T  \right) \\
f_{\bold{V}_j}(\bold{v}_j)=\frac{1}{(2\pi)^{\frac{K}{2}}} \exp\left(-\frac{1}{2}\bold{v}_j\bold{v}_j^T  \right)
\end{cases}
\end{align*}
The likelihood of each observed rating $r_{ij}$ is defined as a normal distribution:
$$
R_{ij}|\bold{U}_i,\bold{V}_j \sim N\left(\text{sigmoid}\left(\bold{u}_i\bold{v}_j^T\right),\sigma^2 \right)
$$
Note: as sigmoid function returns a value into $[0,1]$, during training, we need to first normalize ratings via $r_{ij} \leftarrow \frac{r_{ij}-1}{R-1}$, where original ratings are defined in $\{1, 2, … , R\}$, and $\sigma^2$ is a hyperparameter.

The PDF of likelihood is
\begin{align*}
f_{R_{ij}|\bold{U}_i,\bold{V}_j}(r_{ij}|\bold{u}_i,\bold{v}_j)= \frac{1}{\sqrt{2\pi \sigma^2}}\exp\left( -\frac{\left(r_{ij}-\text{sigmoid}\left(\bold{u}_i\bold{v}_j^T\right)\right)^2}{2\sigma^2} \right).
\end{align*}
Based on conditional independence and Bayes' rule, the posterior of $\bold{U}$ and $\bold{V}$ can be inferred as
\begin{align*}
&f_{\bold{U},\bold{V}|\{\bold{R}_{ij}\}}(U,V|\{r_{ij}\})\\
\propto\ & f_{\{\bold{R}_{ij}\}|\bold{U},\bold{V}}(\{r_{ij}\}|U,V)f_{\bold{U},\bold{V}}(U,V) \\
\propto\ &\prod\limits_{i=1}^{N}\prod\limits_{j=1}^{M} \left[f_{R_{ij}|\bold{U}_i,\bold{V}_j}(r_{ij}|\bold{u}_i,\bold{v}_j)  \right]^{I_{ij}}\exp\left(-\frac{1}{2}\bold{u}_i\bold{u}_i^T  \right)\exp\left(-\frac{1}{2}\bold{v}_j\bold{v}_j^T  \right).
\end{align*}
where $I_{ij}$ is an indicator function that is equal to $1$ if user $i$ rated movie $j$, otherwise $0$.

If the likelihood adopts a simple Gaussian distribution, minimizing the negative log of the posterior (loss function) combined with gradient descent to obtain the most likely $U$ and $V$, and then directly multiplying them to generate the prediction matrix are still available, just as researchers do in PMF \cite{2}. However, we introduce nonlinearity by applying a sigmoid function to the mean of the Gaussian likelihood distribution, in order to simulate the complex likelihoods that are likely to occur in real-world applications. In addition, a simple Gaussian likelihood may produce predicted ratings outside the valid range (e.g., from 1 to 5), while sigmoid can compress the predicted ratings into valid range.

In this case, the gradient descent becomes infeasible. Therefore, we adopt Bayesian framework, retain posterior distributions of $U$ and $V$, and optimizing the predictive distribution over unrated data.

To make a prediction on an unobserved rating $r_{ab}$:
\begin{align*}
&f_{R_{ab}|\{R_{ij}\}}(r_{ab}|\{r_{ij}\})\\
=\ &\int_{\bold{u}_a \in \mathbb{R}^{K}} \int_{\bold{v}_b\in \mathbb{R}^{K}} f_{R_{ab}|\bold{U}_a,\bold{V}_b}(r_{ab}|\bold{u}_a,\bold{v}_b)
f_{\bold{U}_a,\bold{V}_b|\{R_{ij}\}}(\bold{u}_a,\bold{v}_b|\{r_{ij}\})
d\bold{u}_a d\bold{v}_b\\
=\ & \mathbb{E}_{\bold{U}_a,\bold{V}_b|\{R_{ij}\}}\left[ f_{R_{ab}|\bold{U}_a,\bold{V}_b}(r_{ab}|\bold{u}_a,\bold{v}_b)   \right].
\end{align*}
In general, we can use MAP to generate a predicted rating in $[0,1]$:
\begin{align*}
r_{ab}^*=\arg\max\limits_{r_{ab}} f_{R_{ab}|\{R_{ij}\}}(r_{ab}|\{r_{ij}\}).
\end{align*}
Then map back to the original rating using $r=(R-1)r^* +1$.

However, we still cannot compute $ f_{R_{ab}|\{R_{ij}\}}(r_{ab}|\{r_{ij}\}) $, because there are two challenges:

1. Although we have derived the parameter posterior distribution $f_{\bold{U}_a,\bold{V}_b|\{R_{ij}\}}(\bold{u}_a,\bold{v}_b|\{r_{ij}\})$, we cannot accurately compute the normalization constant, since integrating a high-dimensional Gaussian distribution with a sigmoid is difficult even for computers;

2. Even if we obtain the parameter posterior, the integral involved in the predictive distribution itself is high-dimensional and difficult to compute.

To address this, we adopt some Bayesian inference methods.

\section{Bayesian Inference Methods}

Consider Bayesian inference:
\begin{align*}
f_{\Theta|X}(\theta|x)=\frac{f_{\Theta}(\theta)f_{X|\Theta}(x|\theta)}{Z(x)}.
\end{align*}

- \( f_{\Theta|X}(\theta|x) \): Posterior.

- \( f_{\Theta}(\theta) \): Prior.

- \( f_{X|\Theta}(x|\theta) \): Likelihood.

- \( Z(x) \): Normalization constant, where $Z(x)=\int_\theta f_{\Theta}(\theta)f_{X|\Theta}(x|\theta)d\theta$.

For multiple observations, let \( D=\{X_1,\ldots,X_n\} \) be the joint random variables and \( d=\{x_1,\ldots,x_n\} \) their values. Bayesian inference can be written as:
\begin{align*}
f_{\Theta|D}(\theta|d)&=\frac{f_{\Theta}(\theta)f_{D|\Theta}(d|\theta)}{Z(d)}.
\end{align*}
Bayesian prediction:
\begin{align*}
f_{X|D}(x^*|d)
=\int_{-\infty}^{+\infty} f_{X|\Theta}(x^*|\theta)f_{\Theta|D}(\theta|d)d\theta =\mathbb{E}_{\Theta|D=d}[f_{X|\Theta}(x^*|\theta)].
\end{align*}
Two major computational challenges:

1. Computing $Z(d)$ requires computing high dimensional integrals as $\bold{\theta}$ is multivariate in practice;

2. The integral involved in the predictive distribution itself is difficult to compute.

We adopt two solutions:

1. Use MCMC \cite{3,4} to sample from \( f_{\Theta|D}(\theta|d) \) and use the sample mean to approximate the expectation.

2. Use VI \cite{5,6} to approximate \( f_{\Theta|D}(\theta|d) \) by \( q(\theta) \), which is easy to integrate.

\subsection{Markov Chain Monte Carlo (MCMC)}

Suppose $\bold{z}$ is multi-variate random variable, and we are interested in evaluating the expectation:
\begin{align*}
\mathbb{E}_{\bold{Z}} \left[ h(\bold{z})        \right]=\int_{\bold{z}} h(\bold{z})f(\bold{z})d\bold{z}.
\end{align*}
Where
\begin{align*}
f(\bold{z})=\frac{g(\bold{z})}{Z}.
\end{align*}
Our objective is to draw independent samples $\{\bold{z}_1,\ldots, \bold{z}_n\}$ from $f(\bold{z})$ to approximate $\mathbb{E}_{\bold{Z}} \left[ h(\bold{z})        \right]$:
\begin{align*}
\mathbb{E}_{\bold{Z}} \left[ h(\bold{z})        \right] \approx \frac{1}{n} \sum\limits_{i=1}^{n}h(\bold{z}_i).
\end{align*}
In high-dimensional settings, Markov Chain Monte Carlo (MCMC) \cite{3} is widely used for sampling. The Metropolis-Hastings algorithm \cite{3,9} is a commonly used MCMC method.

Core idea: Construct a proposal distribution \( q(\bold{z}'|\bold{z}) \) to generate candidate samples and use an acceptance rate to decide whether to accept the sample.

Detailed steps: Let the target distribution be \( f(\bold{z})=\frac{g(\bold{z})}{Z} \).

1. Construct a proposal distribution \( q(\bold{z}'|\bold{z}) \).

2. Choose an initial state \( \bold{z}_0 \). Set the initial time \( t=0 \).

3. Sample a candidate \( \bold{z}' \) from \( q(\bold{z}'|\bold{z}_t) \).

4. Compute the acceptance rate:
\begin{align*}
\alpha(\bold{z}_t,\bold{z}')= \min \left( 1,\frac{g(\bold{z}')q(\bold{z}_t|\bold{z}')}{g(\bold{z}_t)q(\bold{z}'|\bold{z}_t)}  \right).
\end{align*}
5. Accept the new sample with probability \( \alpha \). If accepted, set \( \bold{z}_{t+1}=\bold{z}' \); otherwise, set \( \bold{z}_{t+1}=\bold{z}_t \).

6. Update time \( t \leftarrow t+1 \).

7. Repeat steps 3–6 until the samples meet the requirements.

The specific iterative process is as follows:

\begin{algorithm}[H]
\caption{Metropolis-Hastings Algorithm for MCMC}
\begin{algorithmic}[1]
\STATE \textbf{Input:} Unnormalized target density $g(\mathbf{z})$, proposal distribution $q(\mathbf{z}'|\mathbf{z})$
\STATE \textbf{Output:} Samples $\{\mathbf{z}_1, \mathbf{z}_2, \ldots, \mathbf{z}_n\}$ approximating $f(\mathbf{z})$
\STATE \textbf{Initialize:} Initial state $\mathbf{z}_0$, set $t = 0$
\WHILE{$t < n$}
    \STATE Sample a candidate $\mathbf{z}' \sim q(\mathbf{z}'|\mathbf{z}_t)$
    \STATE Compute acceptance rate: 
    \[
    \alpha(\mathbf{z}_t, \mathbf{z}') = \min\left(1, \frac{g(\mathbf{z}')q(\mathbf{z}_t|\mathbf{z}')}{g(\mathbf{z}_t)q(\mathbf{z}'|\mathbf{z}_t)}\right)
    \]
    \STATE Sample $u \sim \text{Uniform}(0,1)$
    \IF{$u < \alpha(\mathbf{z}_t, \mathbf{z}')$}
        \STATE Accept: set $\mathbf{z}_{t+1} = \mathbf{z}'$
    \ELSE
        \STATE Reject: set $\mathbf{z}_{t+1} = \mathbf{z}_t$
    \ENDIF
    \STATE Update $t \leftarrow t+1$
\ENDWHILE
\STATE \textbf{return} $\{\mathbf{z}_1, \mathbf{z}_2, \ldots, \mathbf{z}_n\}$
\end{algorithmic}
\end{algorithm}

For mathmatical details, please check Appendix.

\subsection{Variational Inference (VI)}

Due to the intractability of the exact posterior \( P(\mathbf{U}, \mathbf{V} \mid \{r_{ij}\}) \), we employ Variational Inference (VI) \cite{5} to approximate it. We define a variational distribution \( Q(\mathbf{U}, \mathbf{V}) \) under the mean-field assumption:
\begin{align*}
Q(\mathbf{U}, \mathbf{V}) = \prod_{i=1}^N Q_i(\mathbf{u}_i) \prod_{j=1}^M Q_j(\mathbf{v}_j),
\end{align*}
and optimize it to minimize the KL divergence between the true posterior and the variational distribution. This leads to the maximization of the Evidence Lower Bound (ELBO):
\begin{align*}
\log P(\{r_{ij}\}) \geq \mathcal{L}(Q) = \mathbb{E}_{Q}[\log P(\{r_{ij}\}, \mathbf{U}, \mathbf{V})] - \mathbb{E}_{Q}[\log Q(\mathbf{U}, \mathbf{V})].
\end{align*}
We derive the update rules using Coordinate Ascent Variational Inference (CAVI) \cite{10}, iteratively optimizing each variational factor while keeping others fixed.

The specific iterative process is as follows:

\begin{algorithm}
\caption{Coordinate Ascent Variational Inference (CAVI)}
\begin{algorithmic}[1]
\STATE \textbf{Input:} A model $p(\mathbf{x}, \mathbf{z})$, a data set $\mathbf{x}$
\STATE \textbf{Output:} A variational density $Q(\mathbf{z}) = \prod_{j=1}^m Q_j(z_j)$
\STATE \textbf{Initialize:} Variational factors $Q_j(z_j)$
\WHILE{the ELBO has not converged}
    \FOR{$j \in \{1, \dots, m\}$}
        \STATE Set $Q_j(z_j) \propto \exp\left\{ \mathbb{E}_{-j}\left[\log p(z_j \mid \mathbf{z}_{-j}, \mathbf{x})\right] \right\}$
    \ENDFOR
    \STATE Compute $\text{ELBO}(Q) = \mathbb{E}\left[\log p(\mathbf{z}, \mathbf{x})\right] - \mathbb{E}\left[\log Q(\mathbf{z})\right]$
\ENDWHILE
\STATE \textbf{return} $q(\mathbf{z})$
\end{algorithmic}
\end{algorithm}

For mathmatical details, please check Appendix.

\section{Dataset Processing}

We use the MovieLens-small dataset \cite{7}, which consists of 100,836 ratings from 610 users on 9,724 movies. The rating data is stored in the \texttt{ratings.csv} file. We implemented two Python scripts: \texttt{MCMC.py} and \texttt{VI.py}, which perform Bayesian matrix completion using the MCMC and VI methods, respectively, on the data from the CSV file. The source code is available at \href{https://github.com/xx-Weng/Comparing-MCMC-and-VI-for-Bayesian-Probabilistic-Matrix-Factorization-on-the-MovieLens-Dataset}{https://github.com/xx-Weng/Comparing-MCMC-and-VI-for-Bayesian-Probabilistic-Matrix-Factorization-on-the-MovieLens-Dataset}.

The data in \texttt{ratings.csv} is stored in the following format:

\begin{center}
\begin{tabular}{llll}
\toprule
\textbf{userId} & \textbf{movieId} & \textbf{rating} & \textbf{timestamp} \\
\midrule
1 & 1   & 4 & 964982703 \\
1 & 3   & 4 & 964981247 \\
1 & 6   & 4 & 964982224 \\
1 & 47  & 5 & 964983815 \\
1 & 50  & 5 & 964982931 \\
\vdots & \vdots & \vdots & \vdots \\
\bottomrule
\end{tabular}
\end{center}

We ignore the \texttt{timestamp} column and import the first three columns into the Python scripts for further processing. The dataset is preprocessed by:

1. Normalizing ratings between 0 and 1.

2. Divided into three groups: 60\% training set, 20\% validation set, and 20\% test set.


\section{Experimental Evaluation}
We evaluate MCMC and VI on the MovieLens dataset based on:

1. Convergence Speed;

2. Predictive Accuracy;

3. Computational Efficiency.
\subsection{Convergence Speed}
Controlling other parameters such as latent vector dimension and variance to be consistent, we set 300 epochs for VI and found that it generally began to stabilize between 150 and 200 epochs; we set 1000 epochs for MCMC and found that it generally began to stabilize between 600 and 700 epochs. This indicates that VI requires fewer epochs to converge compared to MCMC.

\begin{figure}[H]
    \centering
    \begin{minipage}[t]{0.43\textwidth}
        \centering
        \includegraphics[width=\textwidth]{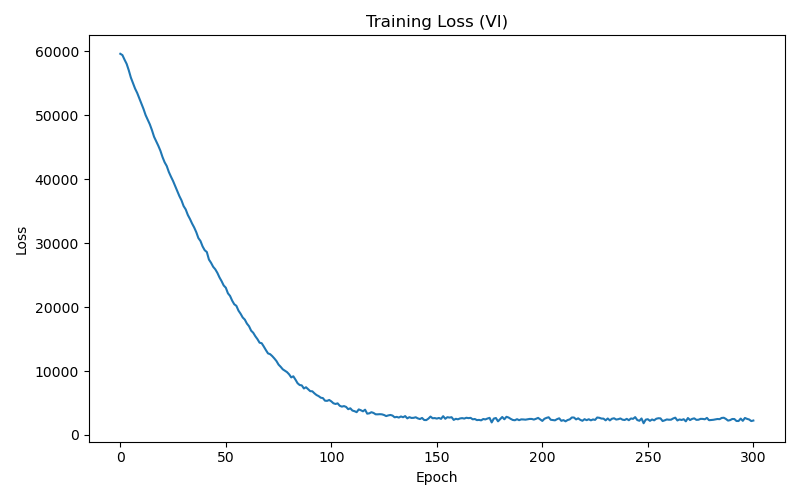}
        \label{fig:example1}
    \end{minipage}
    \hfill
    \begin{minipage}[t]{0.55\textwidth}
        \centering
        \includegraphics[width=\textwidth]{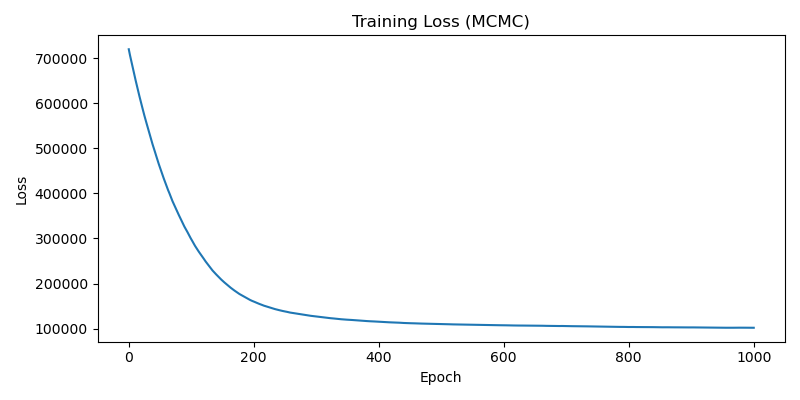}
        \label{fig:example2}
    \end{minipage}
        \caption{Loss-Epoch Plot of VI and MCMC}
\end{figure}

\subsection{Predictive Accuracy}
Measured using RMSE, where VI achieves 1.2277 and MCMC achieves 1.1836.
\begin{figure}[H]
    \centering
    \begin{minipage}[t]{0.49\textwidth}
        \centering
        \includegraphics[width=\textwidth]{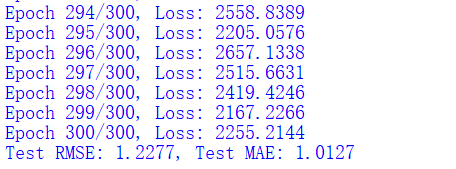}
        \label{fig:example1}
    \end{minipage}
    \hfill
    \begin{minipage}[t]{0.49\textwidth}
        \centering
        \includegraphics[width=\textwidth]{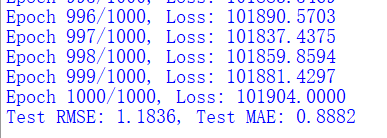}
        \label{fig:example2}
    \end{minipage}
        \caption{RMSE of VI and MCMC}
\end{figure}
\subsection{Computational Efficiency}
We used an NVIDIA GeForce RTX 4060 Laptop GPU for computation. The execution time of VI.py was approximately 6 seconds, while MCMC.py took about 6 hours to run. This means VI runs approximately 3,600x faster than MCMC due to its optimization-based approach.

\subsection{Discussion}

1. MCMC is more accurate as it samples from the true posterior.

2. VI is computationally efficient and suitable for large-scale data.

3. Trade-off: If accuracy is paramount, use MCMC; if speed is critical, use VI.

\section{Conclusion}

In this report, we adopted a Bayesian approach to matrix factorization, leveraging two prominent inference methods—Markov Chain Monte Carlo (MCMC) and Variational Inference (VI)—to address the intractability of the posterior distribution over latent user and item features in collaborative filtering. We formulated the probabilistic model by introducing Gaussian priors on user and item latent vectors and a sigmoid-transformed Gaussian likelihood to ensure bounded rating predictions.

We implemented both inference techniques and evaluated their performance on the MovieLens dataset, focusing on three key dimensions: convergence speed, predictive accuracy, and computational efficiency. Our experiments demonstrated that VI converges considerably faster than MCMC and achieves remarkable computational efficiency due to its deterministic optimization framework. However, MCMC, by virtue of sampling from the true posterior, offers more accurate predictions, though at the cost of significantly higher runtime.

These results highlight a fundamental trade-off in Bayesian matrix factorization: MCMC yields higher fidelity at the expense of time, while VI offers scalability and speed with slightly reduced accuracy. Therefore, the choice of inference method should be guided by the specific constraints and requirements of the target application.

In future work, we plan to explore hybrid approaches that combine the strengths of MCMC and VI, such as initializing MCMC with variational parameters or employing amortized inference techniques. Additionally, extending the model to incorporate content-based features or temporal dynamics could further enhance recommendation accuracy and applicability in real-world systems.

	\clearpage

\begin{center}
\textbf{\huge \\Appendix} 

\end{center}

\textbf{1. Mathmatical Details of MCMC} 

A set of random variables \( \{ \bold{z}_1, \bold{z}_2, \ldots, \bold{z}_n \} \) forms a first-order Markov chain if the following conditional independence holds:
\begin{align*}
P(\bold{z}_{k+1}|\bold{z}_1,\bold{z}_2,\ldots,\bold{z}_k)=P(\bold{z}_{k+1}|\bold{z}_k) \qquad \text{for } k \in \{1,\ldots,n-1\}.
\end{align*}
This means that the current state depends only on the previous state and not on earlier states.

For continuous variables, it becomes:
\begin{align*}
f(\bold{z}_{k+1}|\bold{z}_1,\bold{z}_2,\ldots,\bold{z}_k)=f(\bold{z}_{k+1}|\bold{z}_k) \qquad \text{for } k \in \{1,\ldots,n-1\}.
\end{align*}
To generate a Markov chain, we need:

1. Define the initial state distribution \( P(\bold{z}_1) \).

2. Construct the transition kernels:
\begin{align*}
T_k(\bold{z}_{k+1}\leftarrow\bold{z}_k) = P(\bold{z}_{k+1}|\bold{z}_k),\quad k\in\{1,\ldots,n-1\}.
\end{align*}
For continuous variables, the transition kernel is written as:
\begin{align*}
T_k(\bold{z}_{k+1}\leftarrow\bold{z}_k) = f(\bold{z}_{k+1}|\bold{z}_k),\quad k\in\{1,\ldots,n-1\}.
\end{align*}
A Markov chain is called homogeneous if the transition kernels are the same for all \( k \).

The marginal probability of a specific state can be computed via product and sum rules, i.e., the law of total probability:
\begin{align*}
P(\bold{z}_{k+1})=\sum\limits_{\bold{z}_k}T(\bold{z}_{k+1}\leftarrow \bold{z}_k) P(\bold{z}_k),
\end{align*}
or for continuous variables, expressed in terms of PDFs:
\begin{align*}
f(\bold{z}_{k+1})= \int_{\bold{z}_k} T(\bold{z}_{k+1}\leftarrow \bold{z}_k)f(\bold{z}_k)d\bold{z}_k.
\end{align*}
A distribution \( P^*(\cdot) \) is said to be stationary or invariant with respect to a Markov chain if each step in the chain leaves \( P^*(\cdot) \) invariant:
\begin{align*}
P^*(\bold{z})= \sum\limits_{\bold{z}'} T(\bold{z} \leftarrow \bold{z}')P^*(\bold{z}'), \qquad \forall \bold{z}.
\end{align*}
A sufficient (but not necessary) condition for ensuring \( P^*(\cdot) \) is stationary or invariant is to choose a transition kernel that satisfies the property of detailed balance, defined by:
\begin{align*}
T(\bold{z}' \leftarrow \bold{z})P^*(\bold{z})=T(\bold{z}\leftarrow \bold{z}')P^*(\bold{z}'),\qquad \forall \bold{z},\bold{z}'.
\end{align*}
If detailed balance holds, the chain is said to be reversible.

The core idea of using Markov Chain Monte Carlo (MCMC) methods for sampling:

1. Construct a Markov chain whose stationary distribution is \( f(\bold{z}) \);

2. Simulate the chain to generate samples;

3. After a sufficient number of steps, the samples approximately follow the distribution \( f(\bold{z}) \).

When the Markov chain runs long enough, even if the initial state is not sampled from \( f(\bold{z}) \), the eventually generated samples will converge to this distribution, thus allowing approximate sampling. It should be noted that, to guarantee convergence to \( f(\bold{z}) \), the Markov chain needs to be ergodic.

	\clearpage

\textbf{2. Mathmatical Details of VI} 

Considering the general case, by Bayes' theorem, we have
\begin{align*}
\log P(X) = \log P(X, Z) - \log P(Z \mid X).
\end{align*}
Divide both terms inside the \(\log\) on the right-hand side by \(Q(Z)\). The equation remains equivalent:
\begin{align*}
\log P(X) = \log \frac{P(X, Z)}{Q(Z)} - \log \frac{P(Z \mid X)}{Q(Z)}.
\end{align*}
Multiply both sides of the equation by \(Q(Z)\) and integrate over \(Z\), yielding:

Left-hand side:
\begin{align*}
\int_Z \log P(X) Q(Z) dZ = \log P(X).
\end{align*}
Right-hand side:
\begin{align*}
&\int_Z Q(Z) \log \frac{P(X, Z)}{Q(Z)} dZ - \int_Z Q(Z) \log \frac{P(Z \mid X)}{Q(Z)} dZ\\
=\ &\int_Z Q(Z) \log \frac{P(X, Z)}{Q(Z)} dZ - \int_Z Q(Z) \log P(Z \mid X) dZ\\
=\ &\mathcal{L}(Q) + \mathrm{KL}(Q \| P).
\end{align*}
In the above, the first term is denoted as \(\mathcal{L}(Q)\), and the second term (with a negative sign) represents the KL divergence, indicating the distance between the posterior distribution \(P(Z \mid X)\) and the distribution \(Q(Z)\). Therefore, we have:
\begin{align*}
\log P(X) = \mathcal{L}(Q) + \mathrm{KL}(Q \| P).
\end{align*}
Assume under the mean-field theory that \(Q(Z)\) is conditionally independent for all components of \(Z\) (let \(Z\) have \(M\) components), i.e.,
\begin{align*}
Q(Z) = \prod_{i=1}^M Q_i(Z_i). 
\end{align*}
From the previous derivation, we know:
\begin{align*}
\mathcal{L}(Q) =\ & \int_Z Q(Z) \log \frac{P(X, Z)}{Q(Z)} dZ  \\
=\ & \int_Z \prod_{i=1}^M Q_i(Z_i) \log P(X, Z) dZ - \int_Z \prod_{i=1}^M Q_i(Z_i) \log \prod_{i=1}^M Q_i(Z_i) dZ\\
=\ & \int_Z \prod_{i=1}^M Q_i(Z_i) \log P(X, Z) dZ - \int_Z \prod_{i=1}^M Q_i(Z_i) \sum_{i=1}^M \log Q_i(Z_i) dZ. 
\end{align*}
We will make some transformations to the first expression on the left side:
\begin{align*}
&\int_Z \prod_{i=1}^M Q_i(Z_i) \log P(X, Z) dZ  \\
=\ & \int Q_j(Z_j) \left[ \int \prod_{i \neq j} Q_i(Z_i) \log P(X, Z) dZ_j \right] dZ_j\\
=\ & \int q_j(Z_j) \mathbb{E}_{q_i(Z_i), i \neq j} \left[ \log P(X, Z) \right] dZ_j.
\end{align*}
Now let's make some transformations to the second expression on the right side:
\begin{align*}
& \int_Z \prod_{i=1}^M Q_i(Z_i) \sum_{i=1}^M \log Q_i(Z_i) dZ \\
=\ & \int_Z \left( \prod_{i=1}^{M} Q_i(z_i) \right) \left( \sum_{i=1}^{M} \log Q_i(z_i) \right) dZ\\
=\ & \int_Z \left( \prod_{i=1}^{M} Q_i(z_i) \right) \left[ \log Q_1(z_1) + \log Q_2(z_2) + \cdots + \log Q_M(z_M) \right] dZ.
\end{align*}
Now we attempt to isolate one of the items to discover the pattern:
\begin{align*}
&\int_Z \left( \prod_{i=1}^{M} Q_i(z_i) \right) \log Q_1(z_1) \, dZ\\
=\ & \int_Z Q_1 Q_2 \cdots Q_M \log Q_1 \, dZ\\
=\ & \int_{z_1, z_2, \dots, z_M} Q_1 Q_2 \cdots Q_M \log Q_1 \, dz_1 dz_2 \cdots dz_M\\
=\ & \left( \int_{z_1} Q_1 \log Q_1 \, dz_1 \right) \left( \int_{z_2} Q_2 \, dz_2 \right) \cdots \left( \int_{z_M} Q_M \, dz_M \right)\\
=\ & \int_{z_1} Q_1 \log Q_1 \, dz_1 .
\end{align*}
Back to the expression we care about
\begin{align*}
&\int_Z \left( \prod_{i=1}^{M} Q_i(z_i) \right) \left[ \log Q_1(z_1) + \log Q_2(z_2) + \cdots + \log Q_M(z_M) \right] dZ\\
= \ & \sum_{i=1}^{M} \left( \int_{z_i} Q_i(z_i) \log Q_i(z_i) \, dz_i \right)\\
=\ & \int_{z_j} Q_j(z_j) \log Q_j(z_j) \, dz_j + \text{Constant}.
\end{align*}
We transform the expectation to another form:
\begin{align*}
\mathbb{E}_{Q_i(Z_i), i \neq j} \left[ \log P(X, Z) \right]=\log \tilde{P}(X, Z_j).
\end{align*}
Here we adopt Coordinate Ascent Variational Inference (CAVI), the idea of CAVI is that when updating \( Q_j(Z_j) \), the other \( Q_i(Z_i) \) for \( i \neq j \) are kept fixed, thus
\begin{align*}
&\mathcal{L}(Q) =  \int_{z_j} Q_j(z_j) \log \frac{\hat{P}(X, z_j)}{Q_j(z_j)} \, dz_j + \text{Constant}\\
=\ & - \mathrm{KL}\left( Q_j \, \| \, \hat{P}(X, z_j) \right) + \text{Constant}.
\end{align*}
To maximize \(\mathcal{L}(\mathbf{Q})\), we need to set \(Q_j(Z_j) = \tilde{P}(X, Z_j)\), that is,
\begin{align*}
Q_j^*(Z_j) = \exp\left( \mathbb{E}_{Q_i(Z_i), i \neq j} \left[ \log P(X, Z) \right] \right).
\end{align*}
To ensure that \(\sum_{Z_j} Q_j^*(Z_j) = 1\), we normalize the above expression, obtaining
\begin{align*}
Q_j^*(Z_j) = \frac{ \exp\left( \mathbb{E}_{Q_i(Z_i), i \neq j} \left[ \log P(X, Z) \right] \right) }{ \int \exp\left( \mathbb{E}_{Q_i(Z_i), i \neq j} \left[ \log P(X, Z) \right] \right) dZ_j }.
\end{align*}
In this project, Z can be regarded as [U; V], X can be regarded as $\{r_{ij}\}$
\begin{align*}
P(X, Z) &= f(\mathbf{U}, \mathbf{V} , \{r_{ij}\}) \\
&= \prod_{i=1}^N \prod_{j=1}^M 
\left( \mathcal{N}(r_{ij} \mid \text{sigmoid}(\mathbf{u}_i^\top \mathbf{v}_j), \sigma^2) \right)^{I_{ij}}
\exp\left(-\frac{1}{2} \mathbf{u}_i^\top \mathbf{u}_i \right)
\exp\left(-\frac{1}{2} \mathbf{v}_j^\top \mathbf{v}_j \right).
\end{align*}


\begin{thebibliography}{10}
\bibitem{1}
Koren, Y., Bell, R., \& Volinsky, C. (2009). Matrix factorization techniques for recommender systems. Computer, 42(8), 30-37.

\bibitem{2}
Mnih, A., \& Salakhutdinov, R. R. (2007). Probabilistic matrix factorization. Advances in neural information processing systems, 20.

\bibitem{3}
W. K. Hastings. (1970). Monte Carlo sampling methods using Markov chains and their applications, Biometrika, Volume 57, Issue 1, April 1970, Pages 97–109, https://doi.org/10.1093/biomet/57.1.97

\bibitem{4}
Ruslan Salakhutdinov and Andriy Mnih. (2008). Bayesian probabilistic matrix factorization using Markov chain Monte Carlo. In Proceedings of the 25th international conference on Machine learning (ICML '08). Association for Computing Machinery, New York, NY, USA, 880–887. https://doi.org/10.1145/1390156.1390267

\bibitem{5}
Blei, D. M., Kucukelbir, A., \& McAuliffe, J. D. (2017). Variational Inference: A Review for Statisticians. Journal of the American Statistical Association, 112(518), 859–877. https://doi.org/10.1080/01621459.2017.1285773

\bibitem{6}
Guangyong Chen, Fengyuan Zhu, and Pheng Ann Heng. (2018). Large-Scale Bayesian Probabilistic Matrix Factorization with Memo-Free Distributed Variational Inference. ACM Trans. Knowl. Discov. Data 12, 3, Article 31 (June 2018), 24 pages. https://doi.org/10.1145/3161886

\bibitem{7}
F. Maxwell Harper and Joseph A. Konstan. 2015. The MovieLens Datasets: History and Context. ACM Transactions on Interactive Intelligent Systems (TiiS) 5, 4: 19:1–19:19. https://doi.org/10.1145/2827872

\bibitem{8}
Su, X., \& Khoshgoftaar, T. M. (2009). A survey of collaborative filtering techniques. Advances in artificial intelligence, 2009(1), 421425.



\bibitem{9}
Siddhartha Chib \& Edward Greenberg (1995) Understanding the Metropolis-Hastings Algorithm, The American Statistician, 49:4, 327-335, DOI: 10.1080/00031305.1995.10476177

\bibitem{10}
Lee, S. Y. (2022). Gibbs sampler and coordinate ascent variational inference: A set-theoretical review. Communications in Statistics-Theory and Methods, 51(6), 1549-1568.

\end{thebibliography}
\end{document}